\documentclass[letterpaper]{article} 
\usepackage{aaai23}  
\usepackage{times}  
\usepackage{helvet}  
\usepackage{courier}  
\usepackage[hyphens]{url}  
\usepackage{graphicx} 
\usepackage{natbib}  
\usepackage{caption} 
\usepackage{algorithm}
\usepackage{algorithmic}
\usepackage{amsmath}
\usepackage{amsthm,amssymb}
\usepackage{mathrsfs}
\usepackage{booktabs}
\usepackage{bbding}
\usepackage{color}
\usepackage{multirow}
\usepackage{newfloat}
\usepackage{listings}
\DeclareCaptionStyle{ruled}{labelfont=normalfont,labelsep=colon,strut=off} 
\floatstyle{ruled}
\newfloat{listing}{tb}{lst}{}
\floatname{listing}{Listing}
\title{Dialogue State Distillation Network with Inter-slot Contrastive Learning for Dialogue State Tracking}
\author {
    Jing Xu\textsuperscript{\rm 1}\thanks{Work done during internship at Baidu Inc.},
    Dandan Song\textsuperscript{\rm 1}\thanks{Corresponding authors.},
    Chong Liu\textsuperscript{\rm 2},
    Siu Cheung Hui\textsuperscript{\rm 3},\\
    Fei Li\textsuperscript{\rm 2},
    Qiang Ju\textsuperscript{\rm 2},
    Xiaonan He\textsuperscript{\rm 2},
    Jian Xie\textsuperscript{\rm 2}\footnotemark[2]
}
\affiliations {
    \textsuperscript{\rm 1}School of Computer Science \& Technology, Beijing Institute of Technology, Beijing, China\\
    \textsuperscript{\rm 2}Baidu Inc., Beijing, China\\
    \textsuperscript{\rm 3}Nanyang Technological University, Singapore\\
    \{xujing, sdd\}@bit.edu.cn, \{liuchong06, lifei21, juqiang, hexiaonan, xiejian01\}@baidu.com, asschui@ntu.edu.sg
}

\usepackage{bibentry}

\begin{document}

\maketitle

\begin{abstract}
In task-oriented dialogue systems, Dialogue State Tracking (DST) aims to extract users' intentions from the dialogue history. Currently, most existing approaches suffer from error propagation and are unable to dynamically select relevant information when utilizing previous dialogue states. Moreover, the relations between the updates of different slots provide vital clues for DST. However, the existing approaches rely only on predefined graphs to indirectly capture the relations. In this paper, we propose a Dialogue State Distillation Network (DSDN) to utilize relevant information of previous dialogue states and migrate the gap of utilization between training and testing. Thus, it can dynamically exploit previous dialogue states and avoid introducing error propagation simultaneously. Further, we propose an inter-slot contrastive learning loss to effectively capture the slot co-update relations from dialogue context. Experiments are conducted on the widely used MultiWOZ 2.0 and MultiWOZ 2.1 datasets. The experimental results show that our proposed model achieves the state-of-the-art performance for DST.
\end{abstract}

\section{Introduction}
\begin{figure}[]
\centering
\includegraphics[width=1\columnwidth]{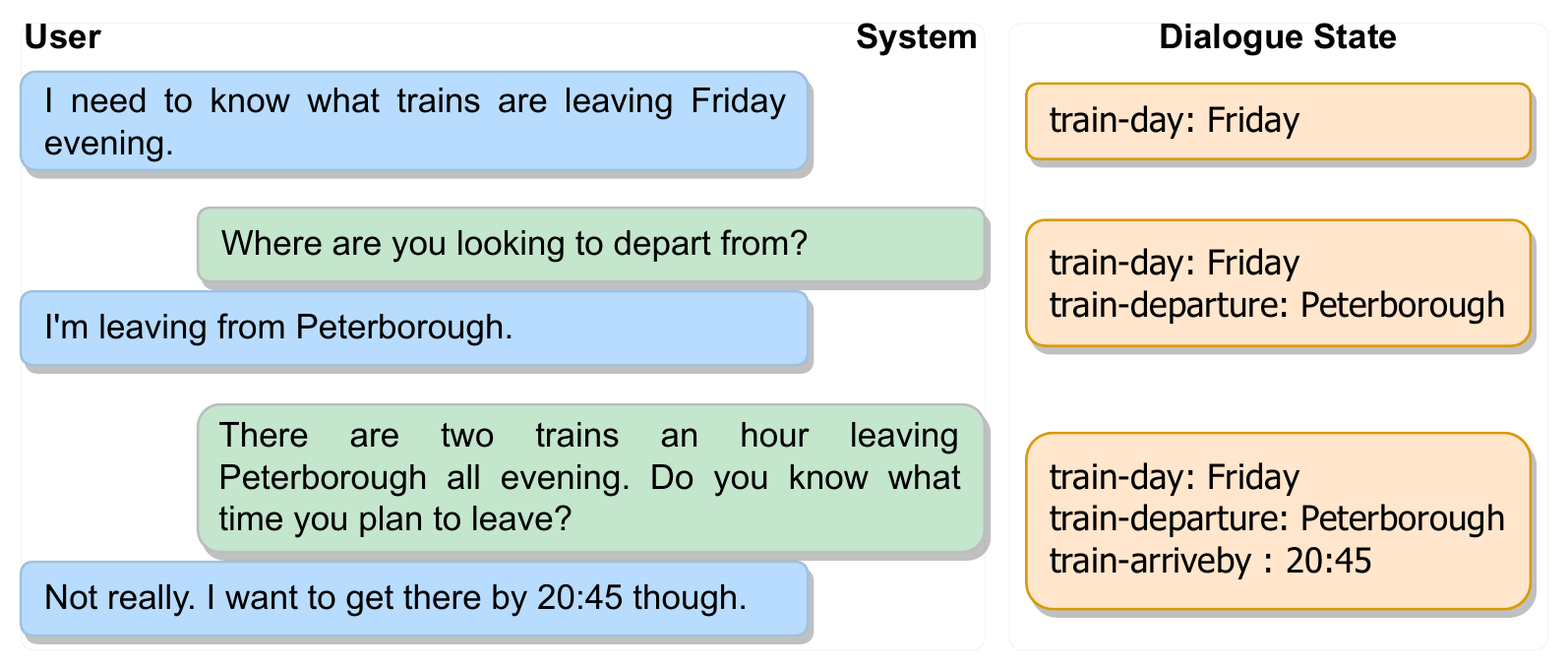} 
\caption{An example of DST. Utterances on the left side are from the user and system over the continuation of the multiple turns of a dialogue. The corresponding dialogue state for each turn is on the right side. Specifically, ``train-day: Friday'' represents that the value of the slot ``day'' in the domain ``train'' is ``Friday''.}
\label{example}
\end{figure}

As a key component of task-oriented dialogue systems, Dialogue State Tracking (DST) aims to extract users' goals or intents over the continuation of multiple turns of a dialogue, which are described as a compact dialogue state. Specifically, a DST model aims to extract the corresponding values of pre-defined slots given a dialogue history. Thus, the extracted dialogue state can be represented as a set of (slot, value) pairs. A DST example is shown in Figure \ref{example}.

As previous dialogue states can provide effective and explicit information of a dialogue history~\cite{kim2020efficient}, recent approaches attempt to utilize the previous dialogue state information for DST. Most of these approaches~\cite{zhu2020efficient, zeng2020jointly, zeng2020multi} concatenate the dialogue context and the corresponding previous dialogue state as input. However, the golden dialogue states can only be obtained during training. Therefore, utilizing the predicted dialogue states may lead to error propagation during the testing process. To alleviate this problem, gated recurrent units are used in a graph attention network to learn to decide which dialogue states are kept~\cite{chen2020schema}. In addition, a uniform scheduled sampling strategy is also used to make the DST model adapt to the data noise of dialogue states, thus improving the robustness of the model~\cite{zhou2021dialogue}. Unfortunately, these attempts are still unable to tackle the problems of the unavailable golden dialogue states and the potential error of predicted dialogue states, though they strive to reduce the negative impact of wrongly predicted dialogue states during the testing process. Moreover, for different current dialogue contexts, the relevant information of previous dialogue states should be dynamically obtained in contrast to missing in the current approaches. Therefore, it is a challenging problem on how to dynamically incorporate previous dialogue states without error propagation for DST. 

Further, we notice that State Operation Prediction (SOP), which aims to predict whether the value of a given slot needs to be updated compared with the previous turn, is an important auxiliary task for DST~\cite{zhu2020efficient,zhang2020find,zhou2021dialogue}. Most existing approaches conduct the prediction of each slot independently. However, slot co-update relations provide a vital clue for the prediction. For example, if users provide their check-in time, they will also tend to provide their hotel stay days when booking a hotel. Thus, the values of the slots ``book day'' and ``book stay'' are usually updated at the same time in the domain ``hotel''. Although several recent approaches incorporate the slot relations across domains to indirectly capture the slot co-update relations for DST~\cite{chen2020schema,zeng2020multi,zhu2020efficient,feng2022dynamic}, they rely on predefined schema graphs or designed graphs and fail to capture the relations from dialogue context automatically. Recent works~\cite{ye2021slot,wang2022luna} also attempt to use self-attention to capture slot dependency. However, they mainly provide mutual guidance among slots in an implicit manner. Therefore, how to explicitly capture the slot co-update relations without predefined graphs is another important issue for DST.     

In this paper, we propose an effective model equipped with a Dialogue State Distillation Network (DSDN) to dynamically exploit previous dialogue states without introducing error propagation. In DSDN, a combination of turn-level and dialogue-level slot relevant information is first encoded by an encoder. Then, a teacher network with golden previous dialogue states tackles SOP while a student network with slot only concatenation learns to imitate the behavior of the teacher network. As a result, the student network can utilize the previous dialogue states without extra input. Moreover, the rich interactions between the encoder and the teacher/student network are conducted to push the teacher/student network to dynamically decide the utilization of previous dialogue states. Finally, the well-learned student network can help our model tackle DST more effectively. 

Additionally, an inter-slot contrastive learning loss is also proposed to explicitly capture the slot co-update relations. In details, we construct the contrastive learning loss between slot representations from both turn-level and dialogue-level according to the labels of SOP. In this way, the model can directly capture the label correlations of SOP between slots automatically and then help improve DST performance. In summary, the main contributions of this paper are as follows:

\begin{itemize}
    \item We propose an effective Dialogue State Distillation Network to dynamically exploit previous dialogue states without introducing error propagation. To the best of our knowledge, this study is the first to exploit knowledge distillation for DST.
    \item We propose an inter-slot contrastive learning loss to
    capture the slot co-update relations from dialogue context automatically. 
    \item Extensive experiments are conducted on the widely used
    MultiWOZ 2.0 and MultiWOZ 2.1 datasets. Our proposed
    model
    achieves the state-of-the-art performance on MultiWOZ 2.1 and competitive performance on MultiWOZ 2.0.
\end{itemize}

\section{Related Work}
\paragraph{Dialogue State Tracking} In task-oriented dialogue systems, as natural language understanding~\cite{wang2022promda,xing2022co} and natural language generation~\cite{mi2022cins} are standalone and less interactive with each other, recent works focus more on dialogue management such as dialogue state tracking~\cite{zhang2020recent}. Existing approaches can be divided into two categories: ontology-based and ontology-free. The ontology-based approaches~\cite{zhong2018global,ramadan2018large,ye2021slot} require an ontology including all possible predefined values for each slot and are simplified into a multi-class classification task. Conversely, the ontology-free approaches~\cite{xu2018end,hosseini2020simple,gao2019dialog,feng2022dynamic} generate the values of slots from dialogue context or vocabularies. In this paper, our work is related to the ontology-based approaches. 

Recently, the utilization of previous dialogue states has attracted much attention for DST. Several works~\cite{kim2020efficient,zhu2020efficient,zeng2020jointly, zeng2020multi} use the previous dialogue states as part of the input. Once a slot in the dialogue states is wrongly predicted, the error information will continue propagating until the end of the dialogue. To improve the utilization effectiveness, the previous dialogue state information is controlled by a recurrent graph attention network~\cite{chen2020schema}. A multi-level fusion gate and a uniform scheduled sampling strategy are also used to reduce the negative impact of wrong predictions~\cite{zhou2021dialogue}.

As for slot co-update relations, most existing approaches~\cite{zhu2020efficient,chen2020schema,wu2020gcdst,ouyang2020dialogue,lin2021knowledge,feng2022dynamic} use graph neural networks based on predefined schema graphs to exploit the prior knowledge of domain or slot dependency, thus capturing them indirectly. Among these models, the co-update relations between slots are only explicitly defined by~\citeauthor{feng2022dynamic}~\shortcite{feng2022dynamic}. They fuse schema graphs and dialogue context to obtain diverse dynamic slot relations which include the slot co-update relations. The obtained slot relations further enhance the final predictions.

\paragraph{Knowledge Distillation} Knowledge distillation~\cite{hinton2015distilling} aims to distill knowledge from teacher networks and transfer it to student networks. Traditionally, it pushes the student networks to imitate the feature representations or probability distributions of the teacher networks. Recently, knowledge distillation has been widely used in natural language processing~\cite{sun2020knowledge,zhang2020distilling,wei2021trigger,zhou2022bert}. In this paper, we propose a novel knowledge distillation network to migrate the gap of utilizing previous dialogue states between the training and testing processes.

\paragraph{Contrastive Learning}
Contrastive learning is first used in computer vision with self-supervised learning~\cite{chen2020simple}, which increases the similarity between the representations of original data samples and differently augmented data samples. Further, supervised contrastive learning~\cite{khosla2020supervised} extends it and generates positive samples from not only data augmentation but also training instances with the same label. In natural language processing, contrastive learning has been widely used in diverse fields, such as pre-trained language model and information extraction~\cite{wang2021cline,wang2021cleve,das2022container,saha2022explanation}. As for DST, we are the first to introduce contrastive learning to exploit the slot co-update relations from the whole dialogue automatically. 

\section{Proposed Model}

For DST, we use $X=\{(U_{1},R_{1}),\dots,(U_{T},R_{T})\}$ to represent a dialogue with $T$ turns and $S=\{s_{1},\dots,s_{J}\}$ to represent a pre-defined slot set, where $U_{t}$ is the user utterance at the $t$-th dialogue turn, $R_{t}$ denotes the corresponding system response and $J$ denotes the total number of pre-defined slots. For each turn $t$, given $X=\{(U_{1},R_{1}),\dots,(U_{t},R_{t})\} (1 \leq t \leq T)$, the goal of DST is to extract the dialogue state which can be represented as $\mathcal{B}=\{(s_{1},v_{1}^{t}),\dots,(s_{J},v_{J}^{t})\}$, where $v_{j}^{t}$ denotes the value of $s_{j}$ for the $t$-th turn. Note that if a slot is not mentioned, its value will be annotated as ``none''. As for any slot, we follow~\cite{wang2022luna} to use a concatenation of its domain name and slot name to represent it, thus integrating both domain information and slot information.

\begin{figure*}[]
\centering
\includegraphics[width=1\linewidth]{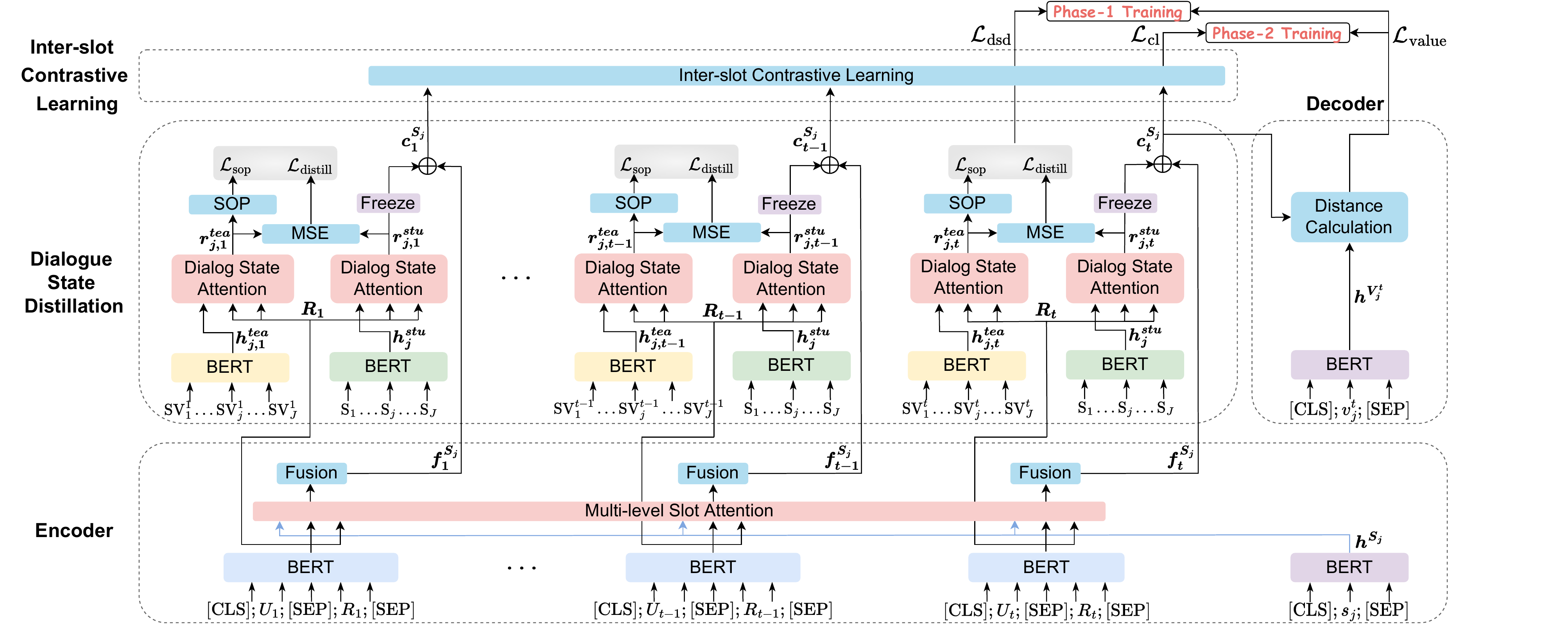}
\caption{The architecture of our proposed model.}
\label{model}
\end{figure*}

In this section, we propose a novel model to tackle DST. Figure \ref{model} shows the proposed model which consists of four modules: Encoder, Dialogue State Distillation, Inter-slot Contrastive Learning and Decoder. Except the Inter-slot Contrastive Learning module, the other three modules constitute the Dialogue State Distillation Network. Next, we will introduce each module in details. 

\subsection{Encoder}
Based on previous works~\cite{lee2019sumbt,ye2021slot,wang2022luna}, we aim to combine turn-level and dialogue-level slot relevant information in this module. First, we use a pre-trained BERT~\cite{kenton2019bert} to encode dialogue context, slots and slot values respectively. Then, we retrieve slot relevant context from dialogue at turn-level and dialogue-level. Finally, a max-pooling operation is used to fuse the retrieved information.

\paragraph{Base Encoder} For the dialogue context at the $t$-th turn, we concatenate the user utterance and the system utterance, and encode them into hidden representations:
\begin{equation}\small
  \boldsymbol{R_{t}}=\textrm{BERT}_{finetune}([\textrm{CLS}]\oplus U_{t} \oplus [\textrm{SEP}] \oplus R_{t} \oplus [\textrm{SEP}])
\label{context}
\end{equation}
where $[\textrm{CLS}]$ and $[\textrm{SEP}]$ denote special tokens in $\textrm{BERT}$, and $\textrm{BERT}_{finetune}$ denotes that the used $\textrm{BERT}$ will be fine-tuned during the training process. For any given slot $s_{j} \in S$ or the corresponding value $v_{j}^{t}$ at the given $t$-th turn, we use another $\textrm{BERT}$ to encode it, whose parameters are fixed during the training process. Unlike encoding dialogue context, we use the hidden representation of the special token $[\textrm{CLS}]$ to represent $s_{j}$ or $v_{j}^{t}$:   
\begin{equation}\small
  \boldsymbol{h^{S_{j}}}=\textrm{BERT}_{fixed}([\textrm{CLS}]\oplus s_{j} \oplus [\textrm{SEP}])
\end{equation}
\begin{equation}\small
  \boldsymbol{h^{V_{j}^{t}}}=\textrm{BERT}_{fixed}([\textrm{CLS}]\oplus v_{j}^{t} \oplus [\textrm{SEP}])
\end{equation}

\paragraph{Multi-level Slot Attention} Based on the hidden representations of dialogue context and slots, we further utilize turn-level and dialogue-level slot attention to obtain slot relevant information. A multi-head attention mechanism~\cite{vaswani2017attention} is first used for the turn-level slot attention and we use $\textrm{MultiHead}(\boldsymbol{Q}, \boldsymbol{K}, \boldsymbol{V})$ to represent it, where $\boldsymbol{Q}$ denotes a query matrix or vector, $\boldsymbol{K}$ and $\boldsymbol{V}$ denote a key matrix and a value matrix respectively. Thus, the turn-level slot attention is denoted as follows: 
\begin{equation}\small
  \boldsymbol{r^{S_{j}}_{t}}=\textrm{MultiHead}(\boldsymbol{h^{S_{j}}}, \boldsymbol{R_{t}}, \boldsymbol{R_{t}})
\end{equation}

As the hidden representations of current dialogue context at the $t$-th turn encoded by Base Encoder lack of previous dialogue context information, we use a transformer encoder~\cite{vaswani2017attention} to aggregate contextual information before dialogue-level slot attention. It is composed of $N$ stacked layers and each layer consists of two sub-layers including a multi-head attention mechanism and a position-wise fully connected feed-forward network. We use the output of the last layer as the contextual representations of each slot, which is denoted as follows:  
\begin{equation}\small
  \boldsymbol{D_{t}^{N}}=\textrm{TransformerEncoder}(\boldsymbol{r^{S_{j}}_{1}},\ldots ,\boldsymbol{r^{S_{j}}_{t}})
\end{equation}

After that, the dialogue-level slot attention is applied to the contextual representations $\boldsymbol{D_{t}^{N}}$, thus summarizing the slot relevant information from the up-to-now utterances for the slot $s_{j}$. Similarly, we use another multi-head attention mechanism as follows:
\begin{equation}\small
  \boldsymbol{d^{S_{j}}_{t}}=\textrm{MultiHead}(\boldsymbol{h^{S_{j}}}, \boldsymbol{D_{t}^{N}}, \boldsymbol{D_{t}^{N}})
\end{equation}

\paragraph{Fusion} To fuse the turn-level and dialogue-level information of slot $s_{j}$, we employ a max-pooling operation to obtain the fusion information $\boldsymbol{f^{S_{j}}_{t}}$ as follows:
\begin{equation}\small
    \left[\boldsymbol{f^{S_{j}}_{t}}\right]_{k}=\max \left\{\left[\boldsymbol{r^{S_{j}}_{t}}\right]_{k},\left[\boldsymbol{d^{S_{j}}_{t}}\right]_{k}\right\}
\end{equation}
where $[\cdot]_{k}$ represents the $k$-th value of a vector.

\subsection{Dialogue State Distillation}
In this module, we aim to learn to distill previous dialogue states from dialogue context without error propagation. To achieve this goal, we first use two BERTs (called the teacher network and the student network respectively) to encode different inputs: golden previous dialogue states and slot only concatenation. Then, we use a multi-head attention mechanism to retrieve relevant information of dialogue states or slots. After that, we train the teacher network to tackle SOP and drive the student network to imitate it simultaneously. Thus, the student network can dynamically utilize relevant information of previous dialogue states without the gap between training and testing. 

\paragraph{Base Encoder} For the teacher network, we concatenate all golden pairs $(s_{j},v_{j}^{t-1},1\leq j \leq J)$ of the $(t-1)$-th turn as the input for the $t$-th turn and use a BERT to encode it:
\begin{equation}\small
  \textrm{SV}_{j}^{t}=[\textrm{SLOT}_{j}^{tea}] \oplus s_{j} \oplus - \oplus v_{j}^{t-1}, 1 \leq j \leq J
\end{equation}
\begin{equation}\small
  \boldsymbol{H^{tea}}=\textrm{BERT}_{finetune}([\textrm{CLS}] \oplus \textrm{SV}_{1}^{t}, \ldots, \textrm{SV}_{J}^{t} \oplus [\textrm{SEP}]) 
\end{equation}
where $[\textrm{SLOT}_{j}^{tea}](1 \leq j \leq J)$ represents special tokens and ``-'' is a separation symbol. Based on $\boldsymbol{H^{tea}}$, we use the representation at the position of $[\textrm{SLOT}_{j}^{tea}](1 \leq j \leq J)$ to represent $(s_{j},v_{j}^{t-1})$, denoted as $\boldsymbol{h^{tea}_{j,t}}$. For the student network, we use slot only concatenation as the input at any turn and use another BERT to encode it:
\begin{equation}\small
  \textrm{S}_{j}=[\textrm{SLOT}_{j}^{stu}] \oplus s_{j}, 1 \leq j \leq J
\end{equation}
\begin{equation}\small
  \boldsymbol{H^{stu}}=\textrm{BERT}_{finetune}([\textrm{CLS}] \oplus \textrm{S}_{1}, \ldots, \textrm{S}_{J} \oplus [\textrm{SEP}]) 
\end{equation}
where $[\textrm{SLOT}_{j}^{stu}](1 \leq j \leq J)$ denotes special tokens. Then, we use the representation at the position of $[\textrm{SLOT}_{j}^{stu}](1 \leq j \leq J)$ to represent $s_{j}$ based on $\boldsymbol{H^{stu}}$, denoted as $\boldsymbol{h^{stu}_{j}}$.

\paragraph{Dialogue State Attention} For the given dialogue context at the $t$-th turn and the slot $s_{j}$, we use $\boldsymbol{h^{tea}_{j,t}}$ and $\boldsymbol{h^{stu}_{j}}$ to retrieve the current dialogue context respectively by two different multi-head attention mechanisms as follows:
\begin{equation}\small
  \boldsymbol{r^{tea}_{j,t}}=\textrm{MultiHead}(\boldsymbol{h^{tea}_{j,t}}, \boldsymbol{R_{t}}, \boldsymbol{R_{t}})
\end{equation}
\begin{equation}\small
  \boldsymbol{r^{stu}_{j,t}}=\textrm{MultiHead}(\boldsymbol{h^{stu}_{j}}, \boldsymbol{R_{t}}, \boldsymbol{R_{t}})
\end{equation}

\paragraph{Distillation} To push the teacher network to hold the slot information of the $t$-th turn, we train it to tackle State Operation Prediction (SOP). SOP aims to identify whether a slot updates its value compared with previous turns. Different from~\cite{zhang2020find,zeng2020jointly}, we follow~\cite{zhou2021dialogue} to simplify it into a binary classification task. In SOP, we use ``1'' to denote that the slot updates its value compared with previous turns, while ``0'' is used if otherwise. We obtain its corresponding loss $\mathcal{L_\textrm{sop}}$ as follows: 
\begin{equation}\small
  p_{j,t}^{sop}=\sigma\left(\boldsymbol{W}^{1} \textrm{tanh}\left(\boldsymbol{W}^{2} \boldsymbol{r^{tea}_{j,t}}\right)\right)
\end{equation}
\begin{equation}\small
  \mathcal{L_\textrm{sop}} =\frac{1}{T\cdot J}\sum_{t=1}^{T} \sum_{j=1}^{J} -(y_{j,t}^{sop} \cdot \log p_{j,t}^{sop} + (1-y_{j,t}^{sop}) \cdot \log (1-p_{j,t}^{sop}))
\end{equation}
where $\sigma$ denotes a sigmoid activation function, $y_{j,t}^{sop}$ denotes the golden label of slot $s_{j}$ for SOP, $\boldsymbol{W}^{1} \in {\mathbb{R}}^{1 \times d_{out}} $ and $\boldsymbol{W}^{2} \in {\mathbb{R}}^{d_{out} \times d_{out}}$ are trainable matrices. Note that $d_{out}$ is the dimension of $\boldsymbol{r^{tea}_{j,t}}$.

Meanwhile, we use a mean square error loss (MSE) to calculate the similarity of the learned representations between the teacher network and the student network for each slot. Thus, the student network can learn to distill previous dialogue state information from the teacher network by minimizing the loss. The distillation loss $\mathcal{L_\textrm{distill}}$ is obtained as follows:
\begin{equation}\small
  \mathcal{L_\textrm{distill}} =\frac{1}{T\cdot J}\sum_{t=1}^{T} \sum_{j=1}^{J} \textrm{MSE}(\boldsymbol{r^{tea}_{j,t}},\boldsymbol{r^{stu}_{j,t}})
\end{equation}

After that, we formulate the overall loss $\mathcal{L_\textrm{dsd}}$ in the module as follows:
\begin{equation}\small
 \mathcal{L_\textrm{dsd}} = \alpha \mathcal{L_\textrm{sop}} + (1-\alpha) \mathcal{L_\textrm{distill}}
\end{equation}
where $\alpha\,(0 < \alpha < 1)$ is a balancing coefficient. 

\subsection{Inter-slot Contrastive Learning}

In this module, we construct contrastive learning between slots to capture the slot co-update relations automatically. As our goal is to capture the co-update relations of slots, we only consider the slots whose labels of SOP are ``1'' in our contrastive learning. Assume that the SOP label of slot $s_{j}$ at the $t$-th turn (denoted as $s_{j}^{t}$) is ``1'' (i.e., $y_{j,t}^{sop}=1$), we introduce the strategy of positive sample selection and negative sample selection, and the calculation of contrastive learning loss as follows: 

\paragraph{Positive Sample Selection} For the slot $s_{j}$ at the $t$-th turn, we select the slots whose labels of SOP are also ``1'' at the same turn as positive samples and the selected set $\mathcal{P}\left(s_{j}^{t}\right)$ is shown as follows:
\begin{equation}\small
  \mathcal{P}\left(s_{j}^{t}\right)=\left\{s_{p}^{t}: (y_{p,t}^{sop}=1) \wedge (p \neq j)\right\}
\end{equation}

As contrastive learning pulls representations of the given instance $s_{j}^{t}$ and the positive samples together, the similarity between the learned representations of slots which update in the same turn will be increased, thus helping the model capture the co-updated relations between different slots.  

\paragraph{Negative Sample Selection} For the slot $s_{j}$ at the $t$-th turn, we select negative samples from both turn-level and dialogue-level. For turn-level, we select the slots whose labels of SOP are ``0'' in the same turn as negative samples. And for dialogue-level, we select the slot $s_{j}$
at the other turns as negative samples. The selected contrastive learning set $\mathcal{I}\left(s_{j}^{t}\right)$ is shown as follows:
\begin{equation}\small
  \mathcal{N}_{turn}\left(s_{j}^{t}\right)=\left\{s_{n}^{t}: (y_{n,t}^{sop}=0) \wedge (n \neq j) \right\}
\end{equation}
\begin{equation}\small
  \mathcal{N}_{dialogue}\left(s_{j}^{t}\right)=\left\{s_{j}^{t'}: t' \neq t\right\}
\end{equation}
\begin{equation}\small
  \mathcal{I}\left(s_{j}^{t}\right)=\mathcal{P}\left(s_{j}^{t}\right) \cup
  \mathcal{N}_{turn}\left(s_{j}^{t}\right) \cup \mathcal{N}_{dialogue}\left(s_{j}^{t}\right)
\end{equation}
where $\mathcal{N}_{turn}\left(s_{j}^{t}\right)$ and $\mathcal{N}_{dialogue}\left(s_{j}^{t}\right)$ denote the turn-level and dialogue-level selected sets respectively. Contrastive learning pushes representations of the given instance $s_{j}^{t}$ and the negative samples in $\mathcal{N}_{turn}\left(s_{j}^{t}\right)$ away. Thus, the similarity of the learned representations between the given instance $s_{j}^{t}$ and the negative samples in $\mathcal{N}_{turn}\left(s_{j}^{t}\right)$ will be reduced. It helps capture the co-updated relations between slots at the same turn. On the other hand, we reduce the similarity of the learned representations between the same slot at different dialogue turns. As a result, distinguishing these subtle representation differences for the same slot can help our model perceive the state changes of each slot better as the dialogue continues. 

\paragraph{Contrastive Learning Loss} We first integrate the output of Encoder and Dialogue State Distillation, thus utilizing the previous dialogue state information:
\begin{equation}\small
    \boldsymbol{c^{S_{j}}_{t}} = \boldsymbol{f^{S_{j}}_{t}}+\boldsymbol{r^{stu}_{j,t}}
\end{equation}

Then, we follow~\cite{chen2020simple, khosla2020supervised} to map $\boldsymbol{c^{S_{j}}_{t}}$ into a new space, where contrastive learning is used to improve the quality of learning, as follows:
\begin{equation}\small
\label{map}
  \boldsymbol{z^{S_{j}}_{t}}=\boldsymbol{W}^{3} \textrm{ReLU}\left(\boldsymbol{W}^{4} \boldsymbol{c^{S_{j}}_{t}}\right)
\end{equation}
where $\boldsymbol{W}^{3} \in {\mathbb{R}}^{d_{1} \times d_{2}}$ and $\boldsymbol{W}^{4} \in {\mathbb{R}}^{d_{2} \times d_{out}}$ are trainable matrices. Note that $d_{1}$ and $d_{2}$ are the dimensional parameters. After that, we obtain the contrastive learning loss $\mathcal{L_\textrm{cl}}$ based on Normalized Temperature-scaled Cross Entropy (NT-Xent)~\cite{chen2020simple} as follows:
\begin{equation}\small
\footnotesize
  {\mathcal{L_\textrm{cl}}}^{j,t}= \sum_{s_{p}^{t} \in \mathcal{P}\left(s_{j}^{t}\right)} \log \frac{\exp \left(\textrm{sim}\left(\boldsymbol{z^{S_{j}}_{t}},\boldsymbol{z^{S_{p}}_{t}}\right) / \tau\right)}{\sum_{s_{n}^{t'} \in \mathcal{I}\left(s_{j}^{t}\right) \backslash s_{j}^{t}}\exp \left(\textrm{sim}\left(\boldsymbol{z^{S_{j}}_{t}},\boldsymbol{z^{S_{n}}_{t'}}\right) / \tau\right)}
\end{equation}
\begin{equation}\small
  \mathcal{L_\textrm{cl}}=\frac{1}{T\cdot J}\sum_{t=1}^{T} \sum_{j=1}^{J} -\frac{1}{|\mathcal{P}\left(s_{j}^{t}\right)|} {\mathcal{L_\textrm{cl}}}^{j,t}
\end{equation}
where $\textrm{sim}\left(\boldsymbol{z^{S_{j}}_{t}},\boldsymbol{z^{S_{p}}_{t}}\right)=\boldsymbol{z^{S_{j}}_{t}}^\top\boldsymbol{z^{S_{p}}_{t}}/\lVert\boldsymbol{z^{S_{j}}_{t}}\rVert_{2} \,\lVert\boldsymbol{z^{S_{p}}_{t}}\rVert_{2}$ and $\tau$ is a temperature parameter.

\subsection{Decoder} In this module, we aim to predict the value for each slot. Given the slot $s_{j}$ at the $t$-th turn, the integrated representation $\boldsymbol{c^{S_{j}}_{t}}$ is first fed into a linear layer with layer normalization as follows:
\begin{equation}\small
  \boldsymbol{o^{S_{j}}_{t}}=\textrm{LayerNorm}\left(\boldsymbol{W}^{5} \boldsymbol{c^{S_{j}}_{t}}\right)
\end{equation}
where $\boldsymbol{W}^{5} \in {\mathbb{R}}^{d_{out} \times d_{out}}$ is a trainable matrix.

After that, following~\cite{ren2018towards,wang2022luna}, we use L2 norm to obtain the distances between $\boldsymbol{o^{S_{j}}_{t}}$ and candidate values of  $s_{j}$. We obtain the loss $\mathcal{L_\textrm{value}}$ as follows:
\begin{equation}\small
  p_{j,t}^{value}=  \frac{\exp\left(-{\lVert \boldsymbol{o^{S_{j}}_{t}}-\boldsymbol{h^{V_{j}^{t}}} \rVert}_{2}\right)}{\sum_{v \in \mathcal{V}\left(s_{j}\right)}\exp \left(-{\lVert \boldsymbol{o^{S_{j}}_{t}}-\boldsymbol{h^{v}} \rVert}_{2}\right)}
\end{equation}
\begin{equation}\small
  \mathcal{L_\textrm{value}} =\frac{1}{T\cdot J}\sum_{t=1}^{T} \sum_{j=1}^{J} - \log p_{j,t}^{value}
\end{equation}
where $\boldsymbol{h^{V_{j}^{t}}}$ is the representation of golden value for $s_{j}$ and $\mathcal{V}(s_{j})$ denotes the candidate value set of $s_{j}$. 

\subsection{Optimization} To push each module to focus on its optimization objective, we employ a two-phase optimization strategy. First, we optimize the joint objective of $\mathcal{L_\textrm{phase1}}$ as follows:
\begin{equation}\small
  \mathcal{L_\textrm{phase1}}=\mathcal{L_\textrm{value}}+\mathcal{L_\textrm{dsd}}
\end{equation}

After that, we fix the parameters of the Dialogue State Distillation module based on development datasets and then optimize the joint objective of $\mathcal{L_\textrm{phase2}}$ as follows:
\begin{equation}\small
  \mathcal{L_\textrm{phase2}}=\mathcal{L_\textrm{value}}+\mathcal{L_\textrm{cl}}
\end{equation}

\section{Performance Evaluation}
 
\subsection{Experimental Setup}
\paragraph{Dataset} We evaluate our proposed model on the widely used MultiWOZ 2.0~\cite{budzianowski2018multiwoz} and MultiWOZ 2.1~\cite{eric2020multiwoz} datasets. MultiWOZ 2.0 includes over 10,000 dialogues and 35 slots across 7 domains. MultiWOZ 2.1 is the modified version of MultiWOZ 2.0 and corrects previous annotation errors of MultiWOZ 2.0.

\paragraph{Metrics} Following previous works on the two datasets, we use Joint Goal Accuracy (Joint GA) as our evaluation metrics. It is calculated as the proportion of dialogue turns for which the value of each slot is correctly predicted. 

\paragraph{Training Details} For all modules, we follow~\cite{wang2022luna,feng2022dynamic} to use the pre-trained BERT-base-uncased model for Base Encoder. The number of heads is set to 4 for the multi-head attention mechanism. In Encoder, the number of stacked layers in the transformer encoder is set to 6. In Dialogue State Distillation, we set the balancing coefficient $\alpha$ to 0.8 and 0.6 for MultiWOZ 2.0 and MultiWOZ 2.1 respectively. In Inter-slot Contrastive Learning,  we set $d_{1}$ to 512, $d_{2}$ to 512 and the temperature parameter $\tau$ to 0.01. In the two-phase optimization strategy, the epoch of the first phase training is set to 100 and it stops early when the validation loss is not decreased for 15
consecutive epochs. As for the second phase training, we set the epoch to 15. Moreover, we set the batch size to 8 and 16 on four NVIDIA Tesla V100 GPUs during the first phase and second phase training respectively. Adam is used as the optimizer with learning rates of $1 \times 10^{-4}$ and $1 \times 10^{-5}$ during the first phase and second phase training respectively, and the warmup proportion is set to 0.1 during all training processes.

\paragraph{Baselines} We use the following models as baselines for comparison with our proposed model: TripPy~\cite{heck2020trippy} combines three copy strategies for slot value predictions; DST-Picklist~\cite{zhang2020find} first predicts slot updating information and then extracts the values of all slots via category classification; SST~\cite{chen2020schema} integrates schema graph and dialogue context, and controls the slot updating information from previous dialogue states by graph attention networks; Transformer-DST~\cite{zeng2020jointly} jointly optimizes slot updating information prediction and slot value prediction for DST; Graph-DST~\cite{zeng2020multi} constructs a dialogue state graph for each turn of dialogues to replace static schema graphs; FPDSC~\cite{zhou2021dialogue} proposes a multi-level fusion of dialogue context and previous dialogue state information for DST; STAR~\cite{ye2021slot} captures slot correlations by a self-attention mechanism; DSGFNet~\cite{feng2022dynamic} fuses dialogue context and slot-domain relations to generate dynamic slot relations for DST; PPTOD~\cite{su2022multi} learns task-oriented dialogue tasks with a multi-task learning style; LUNA~\cite{wang2022luna} aligns each slot with the utterance which is the most relevant to the slot to enhance DST. Besides, we do not compare our proposed model with TripPy+CoCoAug~\cite{li2021coco}, TripPy+SaCLog~\cite{dai2021preview}, DiCoS-DST~\cite{guo2022beyond} and ASSIST~\cite{ye2022assist} which have benefited from extra resources or more powerful pre-trained language models.

\begin{table}[]
\centering
\small
\begin{tabular}{@{}l|ccc@{}}
\toprule
\multirow{2}{*}{Model} & \multicolumn{2}{c}{MultiWOZ} & \multirow{2}{*}{Total}\\ \cmidrule(l){2-3} 
& 2.0   & 2.1  \\  \hline \hline
 TripPy~\shortcite{heck2020trippy}  & 53.51 & 55.32 & 108.83\\ 
 DST-Picklist~\shortcite{zhang2020find}  & 54.39 & 53.30 & 107.69\\
 SST~\shortcite{chen2020schema}  & 51.17 & 55.23 & 106.40\\
 Transformer-DST~\shortcite{zeng2020jointly}  & 54.64 & 55.35 & 109.99\\
 Graph-DST~\shortcite{zeng2020multi}  & 52.78 & 53.85 & 106.63\\
 FPDSC (turn-level)~\shortcite{zhou2021dialogue}  & 55.03 & 57.88 & 112.91\\
 FPDSC (dual-level)~\shortcite{zhou2021dialogue}  & 53.17 & 59.07 & 112.24\\
 STAR~\shortcite{ye2021slot}  & 54.53 & 56.36 & 110.89\\
 DSGFNet~\shortcite{feng2022dynamic}  & - & 56.70 & -\\ 
 PPTOD~\shortcite{su2022multi}  & 53.89 & 57.45 & 111.34\\
 LUNA~\shortcite{wang2022luna}  & \textbf{55.31} & 57.62 & 112.93\\ \midrule
 Our Model  & 55.30 & \textbf{59.25} & \textbf{114.55}\\ \bottomrule
\end{tabular}
\caption{Experimental results in Joint GA (\%) based on MultiWOZ 2.0 and MultiWOZ 2.1.}
\label{overall}
\end{table}

\subsection{Performance Results}
The experimental results in Joint GA are shown in Table \ref{overall}. From the results, we observe that our model achieves the state-of-the-art performance on MultiWOZ 2.1 and almost the same performance as the state-of-the-art model LUNA on MultiWOZ 2.0. Further, we observe that previous baseline models cannot achieve the most competitive performance on both MultiWOZ 2.0 and MultiWOZ 2.1. Therefore, to compare the comprehensive performance among these models, we sum the performance on the two datasets for each model. As can be seen, our model outperforms the latest state-of-the-art model LUNA by 1.62\%. Moreover, only our model and Graph-DST utilize both previous dialogue states and slot relations. However, our model outperforms Graph-DST by 7.92\% in the summed Joint GA. We infer that Graph-DST relies on predicted dialogue states for enhancing input and capturing slot relations, thus the effectiveness of utilization is limited. Besides, as FPDSC employs fusion gates to reduce error information from predicted dialogue states, it achieves superior performance compared with Graph-DST and Transformer-DST, which also utilize previous dialogue states. It demonstrates the importance of reducing error information when utilizing previous dialogue states and our model uses knowledge distillation to tackle it. 

\begin{figure}[]
\centering
\includegraphics[width=1.0\linewidth]{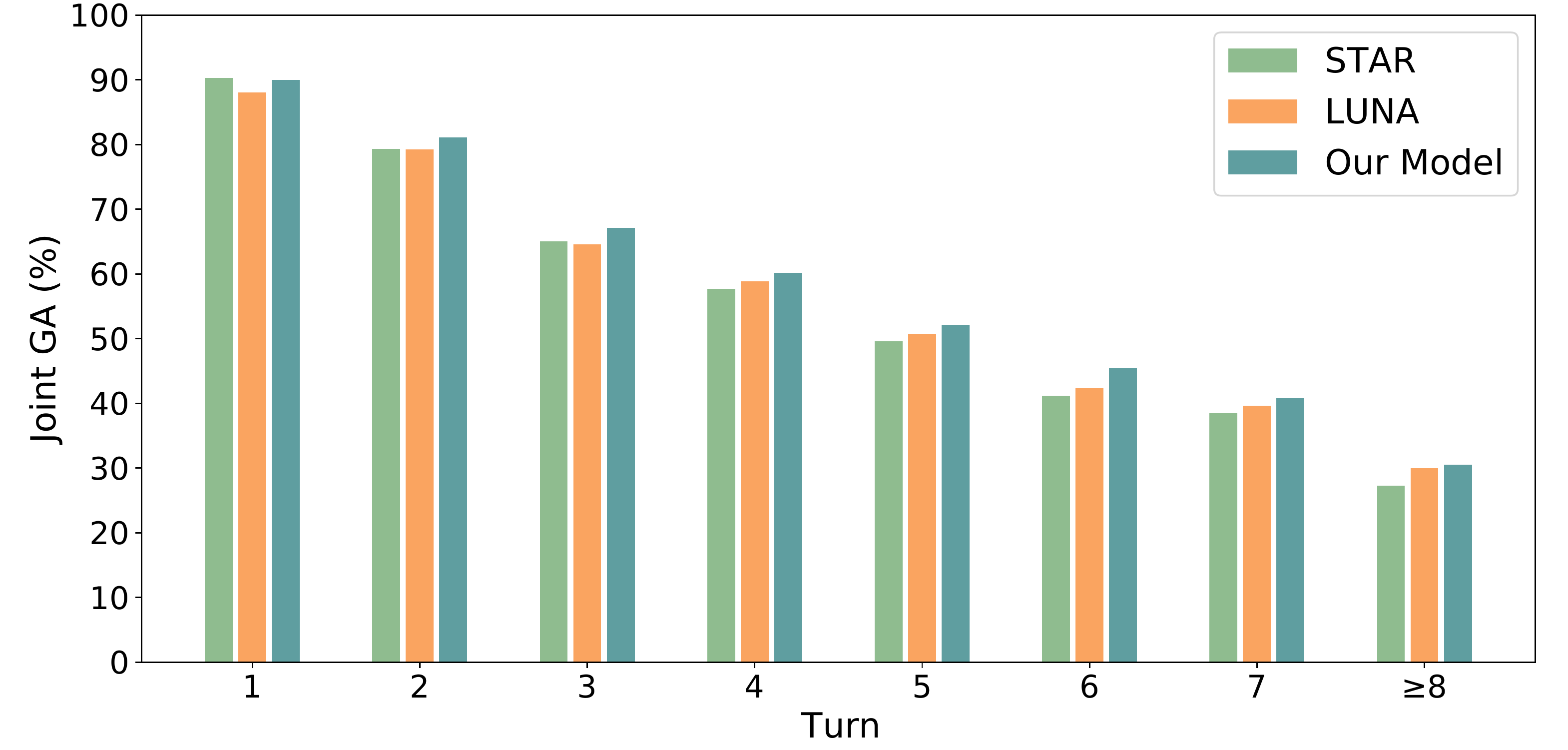}
\caption{Per-turn performance comparison in Joint GA (\%) on MultiWOZ 2.1.}
\label{turn}
\end{figure}

Additionally, we compare the per-turn performance between STAR, LUNA and our model on MultiWOZ 2.1\footnote{As both STAR and LUNA do not present their per-turn performance on MultiWOZ 2.0, we follow them to only present the results on MultiWOZ 2.1 for comparison.}. From the results shown in Figure \ref{turn}, we observe that our model is superior to LUNA and STAR in almost all dialogue turns. It further demonstrates the effectiveness of our model regardless of the length of dialogue.

\begin{table*}[]
\small
\centering
\begin{tabular}{c|l|l|l}
\toprule
\multicolumn{1}{l|}{Turn} & \multicolumn{1}{c|}{Dialogue Context}                                                                                                                                                                                                              & \multicolumn{1}{c|}{Our model w/o Distillation}                                                                                         & \multicolumn{1}{c}{Our model}                                                                                                            \\ \hline
1                         & \begin{tabular}[c]{@{}l@{}}{[}System{]}: There's only one cheap hotel in \\ town, the cambridge belfry, located in the \\ west part of town. Do you need to book a \\ room? \\ {[}User{]}: Does it have a star of 3?\end{tabular} & \begin{tabular}[c]{@{}l@{}}\textbf{hotel-name: cambridge belfry}\\ hotel-pricerange: cheap\\ hotel-star: 3\\ hotel-type: hotel\end{tabular}    & \begin{tabular}[c]{@{}l@{}}\textbf{hotel-name: cambridge belfry}\\ hotel-pricerange: cheap\\ hotel-star: 3\\ hotel-type: hotel\end{tabular} \\ \hline
2                         & \begin{tabular}[c]{@{}l@{}}{[}System{]}: No, it has 4 star rating.\\ {[}User{]}: Do you have any expensive hotels\\ with a 3 star rating?\end{tabular}                                                                         & \begin{tabular}[c]{@{}l@{}}\textbf{hotel-name: cambridge belfry}\\ hotel-pricerange:expensive\\ hotel-star: 3\\ hotel-type: hotel\end{tabular} & \begin{tabular}[c]{@{}l@{}}hotel-name: none\\ hotel-pricerange:expensive\\ hotel-star: 3\\ hotel-type: hotel\end{tabular}          \\ \bottomrule
\end{tabular}
\caption{A case study on the MultiWOZ 2.1 dataset. The wrong predictions of slots are in bold.}
\label{case}
\end{table*}

\subsection{Ablation Study}
\begin{table}[]
\centering
\small
\begin{tabular}{@{}l|cc@{}}
\toprule
\multirow{2}{*}{Model} & \multicolumn{2}{c}{MultiWOZ} \\ \cmidrule(l){2-3} 
& 2.0   & 2.1  \\ \midrule
\textbf{Our model} & 55.30    &  59.25    \\
- Dialogue State Distillation &  53.53 (-1.77)  & 57.28 (-1.97) \\
- Inter-slot Contrastive Learning & 54.25 (-1.05) &  58.59 (-0.66)\\
- above two modules & 51.93 (-3.37) & 56.10  (-3.15)\\ 
\bottomrule
\end{tabular}
\caption{Ablation studies on the MultiWOZ 2.0 and MultiWOZ 2.1 datasets with Joint GA (\%).}
\label{ablation}
\end{table} 

We conduct an ablation study on the MultiWOZ 2.0 and MultiWOZ 2.1 datasets to analyze the effectiveness of each module in our model. As shown in Table \ref{ablation}, we observe that both the Dialogue State Distillation and Inter-slot Contrastive Learning modules contribute significantly to the performance of our proposed model regardless of the used dataset. Here, we focus the discussion on MultiWOZ 2.0. Firstly, if we remove the Dialogue State Distillation and Inter-slot Contrastive Learning modules separately, the performance of the proposed model will be dropped by 1.77\% and 1.05\% respectively in Joint GA. It demonstrates that both the Dialogue State Distillation and Inter-slot Contrastive Learning modules can boost the performance of the proposed model. Secondly, if we train with only $\mathcal{L_\textrm{value}}$, the performance of the proposed model will be decreased significantly by 3.37\% in Joint GA. It further demonstrates the effectiveness of distilling previous dialogue states from dialogue context and capturing the slot co-update relations. 

\subsection{Further Analysis}
In this section, we study the effectiveness of capturing the slot co-updated relations in our proposed model. First, we remove both the Dialogue State Distillation and Inter-slot Contrastive Learning modules, denoted as ``Baseline''. Based on the ``Baseline'', we use different SOP training losses, which include ``Cross Entropy Loss'' and ``$\textrm{Contrastive Learning Loss}^{-}$'', to replace the ``Contrastive Learning Loss'' in the Inter-slot Contrastive Learning module. Note that ``$\textrm{Contrastive Learning Loss}^{-}$'' means that the dialogue-level selected negative sample set $\mathcal{N}_{dialogue}\left(s_{j}^{t}\right)$ is removed from $\mathcal{N}\left(s_{j}^{t}\right)$. Table \ref{sop} shows the performance comparison between the different variant models on the MultiWOZ 2.0 and MultiWOZ 2.1 datasets. As space is limited, we focus the analysis on MultiWOZ 2.0. From the results, we observe that using ``Cross Entropy Loss'' to learn SOP can improve the performance of ``Baseline'' by 0.52\%.  It shows that pushing the model to predict the updating information of slots has achieved some improvements for DST. Moreover, using ``$\textrm{Contrastive Learning Loss}^{-}$'' can improve the performance of ``Baseline'' by 1.22\%.  It indicates that capturing the slot co-updated relations is more effective on the utilization of SOP labels compared with learning to predict the updating information of each slot independently in ``Cross Entropy Loss''. Besides, we observe that using ``Contrastive Learning Loss'' of our model improves the performance of ``Baseline'' by 1.60\%. It demonstrates that the contrastive learning between slots which are from different turns of dialogues can further enhance the performance for DST. 

Moreover, we evaluate the performance difference of SOP when using ``Cross Entropy Loss'' and ``Contrastive Learning Loss'' based on ``Baseline''. For MultiWOZ 2.0, the performance of SOP in Joint GA is 75.52\% when using ``Contrastive Learning Loss'' and outperforms using ``Cross Entropy Loss'' by 2.93\%. As for MultiWOZ 2.1, we obtain the similar performance difference of SOP. It helps present the effectiveness of our contrastive learning more intuitively.  

\subsection{Case Study}
In this section, we discuss a dialogue snippet in the test set of the MultiWOZ 2.1 dataset. Table \ref{case} shows the
prediction results of DST by using ``Our model w/o Distillation'' and our model. ``Our model w/o Distillation'' means that the Dialogue State Distillation module is removed from our model and previous dialogue states are directly concatenated on the dialogue context in Equation (\ref{context}). In the first turn of the dialogue snippet, the system mentions a hotel name ``cambridge belfry'' and the user does not show a clear attitude towards this hotel. Maybe the dialogue context given by the system is too salient, both models ignore the ambiguous attitude of the user and predict the value of ``hotel-name'' wrongly. At the second turn, ``Our model w/o Distillation'' utilizes the wrongly predicted information of ``hotel-name'' at the first turn and the information misleads the prediction of ``Our model w/o Distillation''. As our model uses the Dialogue State Distillation module to distill previous dialogue state information without introducing error propagation, it avoids the disturbance of the wrongly predicted information and predicts the value of ``hotel-name'' as ``none'' correctly.

\begin{table}[]
\centering
\small
\begin{tabular}{@{}l|cc@{}}
\toprule
\multirow{2}{*}{Model} & \multicolumn{2}{c}{MultiWOZ} \\ \cmidrule(l){2-3} 
& 2.0   & 2.1  \\ \midrule
\textbf{Baseline} & 51.93 & 56.10 \\
+ Cross Entropy Loss & 52.45 (+0.52) & 56.76 (+0.66)\\
+ $\textrm{Contrastive Learning Loss}^{-}$ & 53.15 (+1.22) & 57.18 (+1.08)\\
+ $\textrm{Contrastive Learning Loss}$ & 53.53 (+1.60) & 57.28 (+1.18) \\
\bottomrule
\end{tabular}
\caption{Performance comparison in Joint GA (\%) between different
variant models which utilize the label information of SOP on MultiWOZ 2.0 and MultiWOZ 2.1.}
\label{sop}
\end{table}

\section{Conclusion}
In this paper, we propose a Dialogue State Distillation Network (DSDN) to utilize relevant information of previous dialogue states and fill the gap of utilization between training and testing, thus dynamically exploiting previous dialogue states without error propagation. In addition, an inter-slot contrastive learning loss is also proposed to effectively capture the slot co-update relations from dialogue context. Experimental results based on the MultiWOZ 2.0 and MultiWOZ 2.1 datasets have shown that our proposed model has achieved promising performance compared with state-of-the-art models for DST.

\section*{Acknowledgements}
This work is supported by National Key Research and Development Program of China (Grant No.  2020YFC0832606), National Natural Science Foundation of China (Grant Nos. 61976021, U1811262), and Beijing Academy of Artificial Intelligence (BAAI).

\bibliography{aaai23.bib}

\end{document}